\PassOptionsToPackage{table}{xcolor}
\documentclass[sigconf]{acmart}

\renewcommand\footnotetextcopyrightpermission[1]{}
\usepackage{subcaption}
\usepackage[stable]{footmisc}

\AtBeginDocument{%
  \providecommand\BibTeX{{%
    \normalfont B\kern-0.5em{\scshape i\kern-0.25em b}\kern-0.8em\TeX}}}

\acmConference[TextHawk]{}{2024}{ArXiv}

\AtBeginDocument{%
  \providecommand\BibTeX{{%
    \normalfont B\kern-0.5em{\scshape i\kern-0.25em b}\kern-0.8em\TeX}}}

\usepackage{xspace}
\def\modelname{TextHawk\xspace}

\def\bbR{\mathbb{R}}
\def\calB{\mathcal{B}}
\def\calI{\mathcal{I}}
\def\calL{\mathcal{L}}
\def\calM{\mathcal{M}}
\def\calN{\mathcal{N}}
\def\calR{\mathcal{R}}

\def\vare{\boldsymbol{e}}

\def\varg{\boldsymbol{g}}

\def\varv{\boldsymbol{v}}
\def\varx{\boldsymbol{x}}

\makeatletter
\renewcommand\subsubsection{\@startsection{subsubsection}{4}{\z@}%
  {.5em \@plus1ex \@minus.1ex}%
  {-.5em}%
  {\normalfont\normalsize\bfseries}}
\makeatother

\usepackage{tabularx}
\usepackage{multirow}
\usepackage{multicol}
\usepackage{graphicx}

\begin{document}

\title{\modelname: Exploring Efficient Fine-Grained Perception of Multimodal Large Language Models}


\author{Ya-Qi Yu}
\authornote{Both authors contributed equally to this research.}
\affiliation{
    \institution{Huawei Inc.}
    \country{}}
\email{yuyaqi5@huawei.com}

\author{Minghui Liao}
\authornotemark[1]
\affiliation{
    \institution{Huawei Inc.}
    \country{}}
\email{liaominghui1@huawei.com}

\author{Jihao Wu}
\affiliation{
    \institution{Huawei Inc.}
    \country{}}
\email{wujihao@huawei.com}

\author{Yongxin Liao}
\affiliation{
    \institution{Huawei Inc.}
    \country{}}
\email{liaoyongxin@huawei.com}

\author{Xiaoyu Zheng}
\affiliation{
    \institution{Huawei Inc.}
    \country{}}
\email{zhengxiaoyu6@huawei.com}

\author{Wei Zeng}
\affiliation{
    \institution{Huawei Inc.}
    \country{}}
\email{zengwei57@huawei.com}



\begin{abstract}
Multimodal Large Language Models~(MLLMs) have shown impressive results on various multimodal tasks. However, most existing MLLMs are not well suited for document-oriented tasks, which require fine-grained image perception and information compression. In this paper, we present \modelname, a MLLM that is specifically designed for document-oriented tasks, while preserving the general capabilities of MLLMs. \modelname is aimed to explore efficient fine-grained perception by designing four dedicated components. Firstly, a ReSampling and ReArrangement~(ReSA) module is proposed to reduce the redundancy in the document texts and lower the computational cost of the MLLM.  We explore encoding the positions of each local feature by presenting Scalable Positional Embeddings~(SPEs), which can preserve the scalability of various image sizes. A Query Proposal Network~(QPN) is then adopted to initialize the queries dynamically among different sub-images. To further enhance the fine-grained visual perceptual ability of the MLLM, we design a Multi-Level Cross-Attention~(MLCA) mechanism that captures the hierarchical structure and semantic relations of document images. Furthermore, we create a new instruction-tuning dataset for document-oriented tasks by enriching the multimodal document data with Gemini Pro. We conduct extensive experiments on both general and document-oriented MLLM benchmarks, and show that \modelname outperforms the state-of-the-art methods, demonstrating its effectiveness and superiority in fine-grained document perception and general abilities. Project page: \url{https://github.com/yuyq96/TextHawk}.
\end{abstract}






\maketitle

\begin{figure}
    \centering
    \includegraphics[width=\linewidth]{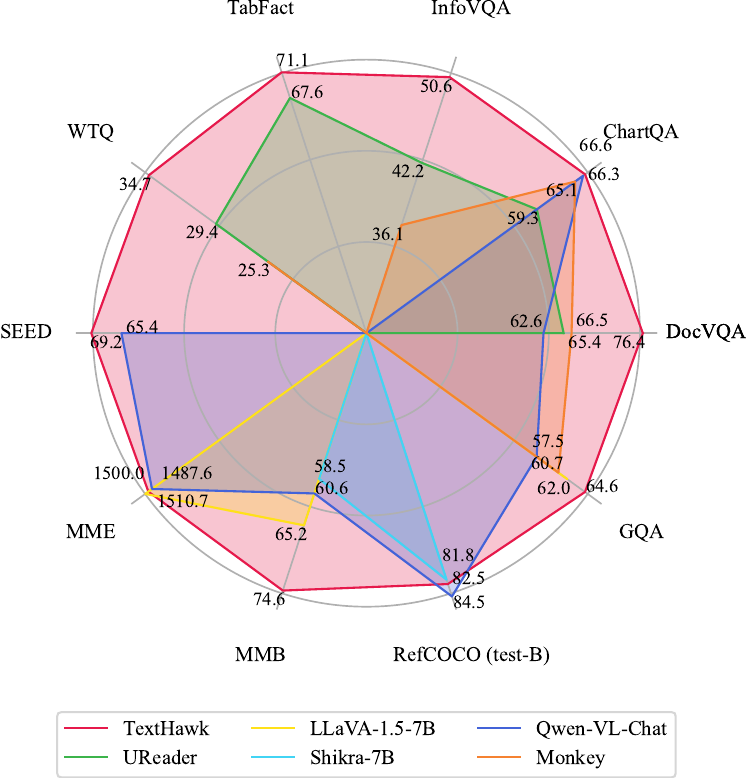}
    \caption{The results of MLLMs on general and document-oriented benchmarks. Best viewed in colors.}
    \label{fig:radar}
\end{figure}

\begin{figure}
    \centering
    \includegraphics[width=\linewidth]{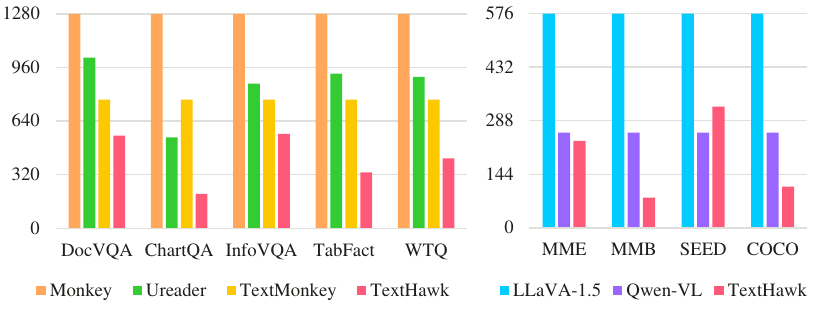}
    \caption{The mean count of compressed visual tokens per image in MLLMs. Best viewed in colors.}
    \label{fig:tokens}
\end{figure}

\section{Introduction}
\label{sec:intro}
Multimodal Large Language Models (MLLMs)~\cite{blip2,instructblip,llava} have received a lot of attention and made great progress recently. They use Large Language Models~(LLMs) as the core and extend the powerful capabilities of LLMs to other modalities, such as visual modalities. Thanks to the wide range of application scenarios of document image understanding, it has a pivotal position in the field of visual perception. Document image understanding ability as one of the core abilities of MLLMs, makes more cutting-edge applications easy to achieve, such as MLLM-based smartphone application agents, rich text-assisted reading, etc. However, document images pose unique challenges for MLLMs, as they differ from natural images in several aspects. Document images typically have higher resolution and higher information density than natural images, which means that MLLMs need to overcome two key difficulties when processing them. The first difficulty is to achieve strong fine-grained visual perception of the document content. The second difficulty is to compress document image information efficiently.

Previous works on document-oriented MLLMs have attempted to solve the difficulties mentioned above. To achieve stronger fine-grained visual perception abilities, Qwen-VL~\cite{qwen-vl} increased the input resolution of the vision encoder from $224\times224$ to $448\times448$ and UReader~\cite{ureader} introduced a shape-adaptive cropping module. To compress the document information, mPLUG-DocOwl~\cite{mplugdocowl} employed a visual abstractor and Qwen-VL utilized a vision-language adapter. These well-designed methods significantly advanced the development of document-oriented MLLMs. Nevertheless, there is still room for further exploration and improvement in fine-grained visual perception and document information compression. Besides, most of the current MLLMs find it difficult to balance both general and document capabilities. Specifically, general MLLMs usually do not focus on improving visual fine-grained perception and information compression, while document-oriented MLLMs may sacrifice general capabilities in their design.

In this paper, we propose \modelname, a multimodal large model that excels at complex document tasks and demonstrates outstanding general capabilities across vision and language domains, as shown in Fig.~\ref{fig:radar}. Considering that simply enlarging the input size of the images can not fit the diverse resolutions of the document images, we follow Ureader~\cite{ureader} to crop the images into sub-images adaptively according to the image shapes.  Based on this, we devise a ReSampling and ReArrangement~(ReSA) module that compresses and rearranges the visual information, which greatly reduces the number of visual tokens, as shown in Fig~\ref{fig:tokens}.  Due to the introduction of the sub-images, we propose Scalable Positional Embeddings~(SPEs) to encode the positions of sub-images while maintaining the scalability across different image sizes. Considering the differences among the sub-images, a Query Proposal Network~(QPN) is then adopted to initialize the queries dynamically among local features. Moreover,  we introduce a Multi-Level Cross-Attention~(MLCA) module that leverages the hierarchical structure and semantic relations of document images to enhance the fine-grained visual perceptual capability. This enables our vision encoder to extract detailed information from dense document images. In addition, we enrich the multimodal document data with Gemini Pro, a commercial MLLM engine, to mitigate the problem of insufficient instruction tuning data.

We address the challenges of fine-grained visual perception and visual information compression for document-oriented MLLMs and propose a new MLLM, named \modelname, that can handle both document-oriented tasks and general vision-language tasks with high performance. The contributions of this paper are as follows:
\begin{enumerate}

    \item We design the ReSA to compress the visual information which significantly reduces the number of visual tokens.

    \item We propose the SPEs and the QPN to fit sub-image representations and enhance the model's fine-grained perception.

    \item We introduce the MLCA that can improve the fine-grained visual perception ability by capturing the global and local information and leveraging the hierarchical structure.

    \item We enrich the multimodal instruction-tuning data of different document-oriented tasks with Gemini Pro. These data can facilitate the fine-tuning of \modelname and benefit the research community.

    \item We demonstrate that \modelname achieves state-of-the-art results on both document benchmarks and general benchmarks, showing its superior fine-grained visual perception and general vision-language abilities.
\end{enumerate}

\section{Related Works}

\subsection{MLLM}
Multimodal Large Language Models (MLLMs) are a class of models that can process and generate multimodal information, mainly including natural language and visual information. They have been shown to achieve remarkable performance on various tasks, such as image captioning, visual question answering,  and visual dialog. Current MLLMs usually consist of a vision encoder, a vision-language adapter, and a large language model. 

BLIP-2~\cite{blip2} proposed a querying transformer (Q-Former) to bridge the frozen image encoder and the frozen large language model. It first learned vision-language representation from a frozen image encoder and then applied vision-to-language generative learning from a frozen language model. InstructBLIP~\cite{instructblip} performed vision-language instruction tuning based on the pre-trained BLIP-2 by introducing an instruction-aware query transformer. LLaVA~\cite{llava} followed a similar architecture while employing a simple linear layer to connect vision and language. It converted image-text pairs into an instruct-following format with ChatGPT/GPT-4 for better fine-tuning results. MiniGPT-4~\cite{minigpt4} adopted a frozen Q-former and a single linear projection layer to align the visual modal and the language modal. LLaVA1.5~\cite{llava-1.5} is an improved version of LLaVA, which adopted a vision encoder with larger input images and a two-layer MLP to improve performance. mPLUG-Owl~\cite{mplugowl} proposed a new training paradigm that enabled the vision encoder and visual abstractor training in the pre-training stage and enabled LoRA with LLM in the instruction tuning stage. mPLUG-Owl2~\cite{mplugowl2} further designed a modality-adaptive module based on mPLUG-Owl and enabled all modules for training. Qwen-VL~\cite{qwen-vl} employed a three-stage training pipeline, including pre-training with image-text pairs, multi-task pre-training with multi-task and interleaved data, and supervised fine-tuning with chat interleaved VL data. 

These methods can understand text images to some extent, but they have limited visual perception for dense documents, especially those with high-resolution images.

\begin{figure*}
    \centering
    \includegraphics[width=\textwidth]{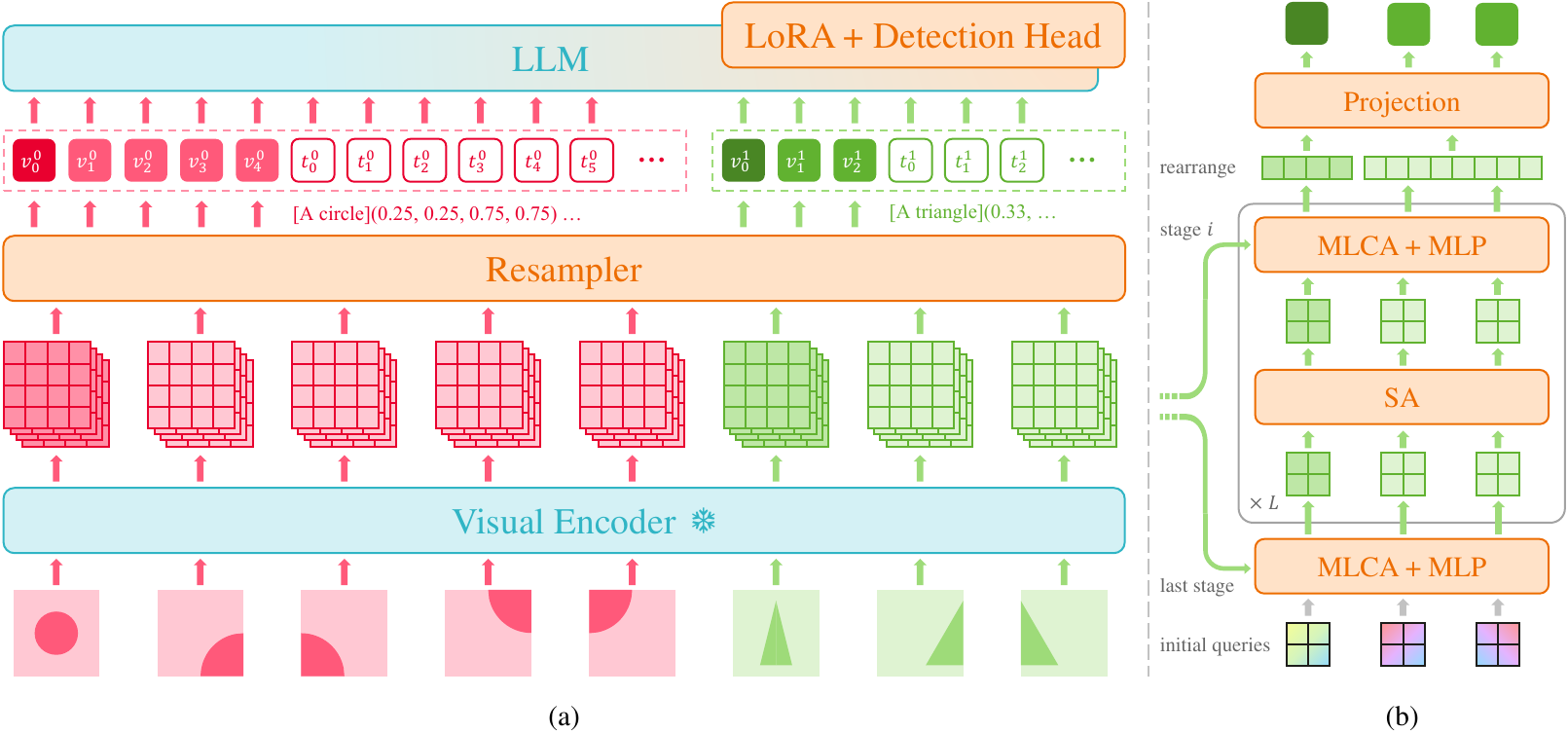}
    \caption{The network architecture and dataflow of \modelname. (a) Overview of \modelname. The visual encoder is frozen throughout the whole training process. (b) Breakdown of \modelname resampler. Image features from different stages are routed to different resampling layers. The text tokenizer, layer normalization, and skip connections are omitted for simplicity.}
    \label{fig:arch}
\end{figure*}

\subsection{Document-Oriented MLLM}
Document-oriented MLLMs are multimodal large language models that can understand text from various types of documents, such as charts, tables, web pages, and scientific papers. They usually incorporate some specific adaptations for document images based on general MLLMs.

mPLUG-DocOwl~\cite{mplugdocowl} followed the mPLUG-Owl model and added some document instruction tuning data, including document, table, webpage, and chart. UReader~\cite{ureader} proposed a shape-adaptive cropping module to obtain better fine-grained visual perceptual ability of document images, based on the pre-trained mPLUG-Owl model. UniDoc~\cite{unidoc} was equipped with text detection and text recognition tasks in its instruction tuning to enhance the ability of text understanding. Monkey~\cite{monkey}, a MLLM with special designs for document images, supported larger resolutions and introduced multi-level description data based on the pre-trained Qwen-VL model. 

Current document-oriented MLLMs mainly focus on adaptation to higher image resolutions and leveraging more document-specific fine-tuning data. Our proposed \modelname also concentrates on the fine-grained visual perception of high-resolution document images and the document data generation, with our novel designs. Moreover, we pay attention to the information compression and the preservation of the general capabilities.

\section{Method}
Our model is designed with two objectives: to effectively process visual inputs of varying resolutions and to compress visual tokens.
 
\subsection{Architecture}
\label{ssec:arch}
The architecture of \modelname is depicted in Fig.~\ref{fig:arch} (a). It consists of a frozen visual encoder, a resampler, and a large language model with a LoRA and a detection head.

\subsubsection{Visual Encoder.} To accelerate image encoding, we prefer a relatively lightweight visual encoder instead of a giant or enormous model. SigLIP~\cite{siglip}, a variant of CLIP~\cite{clip} which adopts Sigmoid loss for vision-language pre-training instead of contrastive learning with Softmax normalization, achieves better zero-shot accuracy in multiple tasks than its competitors. Hence, we employ the Vision Transformer~(ViT) from the efficient SigLIP-SO model as our visual encoder for demonstration, which has different transformer layer configurations but a similar computational cost to the standard ViT-L model. However, all kinds of visual encoders should be feasible in our framework, including models pre-trained in different styles or built with different architectures.

\subsubsection{Resampler.} Similar to Q-Former~\cite{blip2}, our visual token resampler mostly consists of a non-causal transformer decoder which adopts a group of learnable weights as the initial queries and naturally reduces the length of visual features multiple times. For the sake of architecture flexibility, we randomly initialize the resampler instead of initializing it from a pre-trained BERT model or existing resamplers of other MLLMs. Intuitively, we keep the hidden dimension of the intermediate resampler layers equal to that of the visual encoder layers. The resampler has 8 layers and self-attention is removed in the first layer. In order to enhance the awareness of position information during cross-attention, we employ sinusoidal positional encodings and learned positional embeddings for the visual encoder output and the queries respectively at every cross-attention layer.

\subsubsection{Large Language Model.} To facilitate pre-training and take advantage of the interleaved vision-language training, we initialize our 7B LLM with the weights of InternLM-XComposer~\cite{xcomposer}. Similar to BLIP-2, InternLM-XComposer adopts a visual token resampler named perceive sampler to bridge the visual encoder and LLM, but it is anchored on another multi-lingual LLM named InternLM~\cite{internlm}. The architecture of InternLM is almost the same as LLaMA~\cite{llama} except for keeping biases in the attention modules. Specifically, InternLM-XComposer is trained in a two-stage style: The first stage is vision-language pre-training, which incorporates image-text pairs as well as interleaved image-text data. Both the perceived sampler and the LLM are updated in this stage. The second stage is multi-task supervised fine-tuning, in which only the perceived sampler and the LoRA modules are updated. To avoid potential data leakage from the fine-tuning datasets of InternLM-XComposer, we only keep the weights of the LLM from the first pre-training stage and drop all the weights from the vision encoder, perceive sampler, and LoRA modules.

\subsection{Efficient Fine-Grained Perception}
\label{ssec:efgp}

\subsubsection{Shape-Adaptive Cropping.}
The pre-trained visual encoder standardizes image resolution to a fixed and lower size, disregarding the original aspect ratio.
Such processing diminishes the ability to perceive fine-grained content in high-resolution images and introduces notable distortions in aspect ratio.
Following~\cite{ureader}, we augment the frozen ViT by incorporating a dynamic cropping strategy, enabling effective handling of images with arbitrary aspect ratios and resolutions. 
Specifically, an input image $\varv$ with shape $(h\times w)$ will be cropped into multiple sub-images to align with one of the predefined grids $\{\varg=(r\times c)|r,c\in\{1,2,\dots,l\},r\cdot c\leq n\}$, where $r$ and $c$ denotes the rows and columns of the grid $\varg$, $l$ denotes the maximum \emph{side-length}~(number of sub-images in one row or column), and $n$ denotes the maximum \emph{area}~(number of sub-images in the whole image). The grid alignment is regulated by both regular and shape-oriented Intersection over Union~(IoU) measures.
Let us denote the image box as $\text{box}(\varv)=(0,0,h,w)$, the grid box as $\text{box}(\varg)=(0,0,rH,cW)$, and the shape-oriented box as $\text{box}_\text{s}(\varv,\varg)=(0,0,\frac{wr}{h}H,cW)$, where $(H\times W)$ is the input shape of ViT. The IoU values are defined as:
\begin{equation}
    \begin{aligned}
        S_\text{r}(\varv,\varg)&=\text{IoU}(\text{box}(\varv),\text{box}(\varg)),\\
        S_\text{s}(\varv,\varg)&=\text{IoU}(\text{box}_\text{s}(\varv,\varg),\text{box}(\varg)),\\
        S(\varv,\varg)&=S_\text{r}(\varv,\varg)+S_\text{s}(\varv,\varg).
    \end{aligned}
\end{equation}
We select the final grid with the highest summed IoU value $S$, from the top $k$ grids with the highest regular IoU values $S_\text{r}$.

\subsubsection{ReSampling and ReArrangement~(ReSA).} Upon enabling the visual encoder to accept variable resolution input, the number of image tokens can grow exponentially with the image resolution. Without token compression, the maximum number of tokens for a single image reaches $nHW/p^2$ given patch size $p$.
In specific terms, a standard document image aligned with a $5\times4$ grid will consume up to 5,120 tokens.
Previous open-source MLLMs with fine-grained perception capability usually exhibit an image token compression ratio of 4. For instance, Qwen-VL and Monkey reduce the number of image tokens from 1,024 to 256 for each $448\times448$ sub-image, while UReader compresses it from 256 to 64 for each $224\times224$ sub-image. In this case, the consumption of image tokens is still significant. To further explore the possibility of a higher compression ratio, we propose a method combining the advantages of resampling and rearrangement, named ReSA. As shown in Fig.~\ref{fig:arch}~(b), similar to previous MLLMs, ReSA first resamples the image features with a cross-attention mechanism. The hidden dimension of the cross-attention output mirrors that of the visual encoder output, typically being several times smaller than the hidden dimension of the LLMs. Capitalizing on this characteristic, we introduce an additional rearrangement step to further condense the number of image tokens. Following resampling, multiple resampled tokens are concatenated into a single token and then transformed into the latent space of LLMs through a linear projection.
In our experiments, each step of ReSA possesses a compression ratio of 4, resulting in a notably higher compression ratio of 16.

\subsubsection{Multi-Level Cross-Attention~(MLCA).} As mentioned in previous works~\cite{blip2,llava}, visual encoders are pre-trained on specific tasks thus the features from their last layers may focus more on those tasks. It has been proven that the features from the second last layer yield better performance than the last layer~\cite{llava}. Moreover, it is possible to merge features from multiple layers. In the field of object detection, Feature Pyramid Network~(FPN)~\cite{fpn} is well known for merging multi-level features, which improves perception capability on fine-grained objects. As for MLLMs, COMM~\cite{comm} has proved that merging deep and shallow features is beneficial for reducing hallucination and improving performance on fine-grained tasks, even when there is no pyramid structure. Drawing inspiration from FPN, we propose a multi-level feature merging strategy named MLCA. As shown in Fig.~\ref{fig:arch}~(b), MLCA enables the resampler to absorb features from deep as well as shallow visual encoder layers with a predefined routing table. As long as the total number of resampler layers is not changed, MLCA has no extra computational cost compared to the standard cross-attention. Empirically, we adopt 4 visual encoder stages, extracting features from the 14th, 18th, 22nd, and 26th encoder layers respectively.

\begin{figure}[t]
    \centering
    \includegraphics[width=\linewidth]{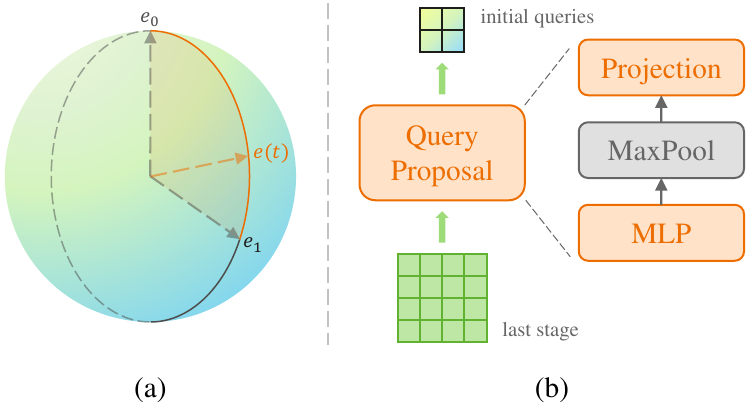}
    \caption{Illustration of (a) scalable positional embeddings interpolation and (b) query proposal network.}
    \label{fig:spe_qpn}
\end{figure}

\subsubsection{Scalable Positional Embeddings~(SPEs).} The relative positional relations among sub-images are ambiguous without the inclusion of additional positional embeddings.
To handle a variable number of image patches, previous works~\cite{pix2struct,ureader} proposed to learn 2-D or factorized absolute positional embeddings covering the maximum positional index presented in the training data. Not only do they lack effectiveness in extrapolation to out-of-domain shapes, but certainly learned embeddings also exhibit under-fitting due to the non-uniform distribution of training input shapes. To overcome the aforementioned obstacles, we propose a novel method named SPEs, extending \emph{factorized} (where row and column are decomposed) positional embeddings to arbitrary shapes. To be clear, the row and column embeddings are handled in the same manner in SPEs, hence their specification is omitted in the following part.

Assume the learned positional embeddings are initialized from a normal distribution $\calN(0, 1)$. Each positional embedding $\vare\in\bbR^d$ is a vector with $\ell_2$-norm $\sqrt{d}$, indicating that the positional embeddings are distributed across the surface of a hypersphere. In practice, the $\ell_2$-norm of learned positional embeddings typically remains within a narrow range during the whole training process, preserving the hypersphere distribution characteristics. Spherical linear interpolation~(Slerp), a commonly employed technique in computer graphics, interpolates any intermediate vector between two unit vectors, emerging as a potential alternative to conventional interpolation methods for positional embeddings.

To strictly meet the requirement of Slerp, we apply normalization and scaling before interpolation for each attention \textit{head}, ensuring uniform $\ell_2$-norm across all positional embeddings:
\begin{align}
    \vare_i&=s\frac{\tilde{\vare}_i}{\|\tilde{\vare}_i\|},
\end{align}
where $\tilde{\vare}_i$~$(i\in\{0,1\})$ denotes two learnable endpoint positional embeddings, and $s$ is a learnable scaling factor initialized as $\sqrt{d}$.

As shown in Fig~\ref{fig:spe_qpn}~(a), we employ Slerp to generate arbitrary positional embeddings spanning between the endpoints:
\begin{equation}
\begin{aligned}
    \theta&=\arccos\frac{\vare_0\vare_1}{\|\vare_0\|\|\vare_1\|},\\
    \vare(t)&=\frac{\sin(\theta-t\theta)}{\sin\theta}\vare_0+\frac{\sin(t\theta)}{\sin\theta}\vare_1,
\end{aligned}
\end{equation}
where $t\in[0,1]$ is the fractional position, which can be the relative position of a sub-image or an image patch.

\subsubsection{Query Proposal Network~(QPN).} Despite the satisfactory performance of Q-Former observed on fixed resolution MLLMs, the way of initializing the resampler queries from a fixed number of learned parameters lacks flexibility under the variable resolution settings. Reusing the initial queries on different sub-images might lead to redundancy and undesired attention patterns, wherein resampled image tokens corresponding to distinct sub-images but identical resampler queries exhibit strong similarities and receive improperly higher attention scores. To eliminate the side-effect of shared initial queries, we propose a lightweight module called QPN for generating the queries dynamically. As shown in Fig~\ref{fig:spe_qpn}~(b), the structure of QPN consists of a 2-layer MLP with GELU activation, a max pooling layer, and a linear projection layer. The output of the visual encoder is fed into QPN and the number of proposed queries is hereby controlled by the stride of the max pooling layer. For a fair comparison, our experiments adopt a stride of $2\times2$ so that the compression ratio remains 4. The output dimension of the MLP layers and the input dimension of the projection layer are set to 4 times the hidden dimension of the visual encoder.

\subsubsection{Detection Head.} Previous works~\cite{shikra,qwen-vl,llava-1.5} on applying MLLMs for localizing target objects mostly adopt plain text for representing coordinates, which is intuitive since pre-trained LLMs work well on regular text strings. However, plain text-based coordinates are token-consuming, lowering both the training throughput and inference efficiency. We propose to expand the vocab of MLLMs with special tokens for normalized coordinates. Specifically, employing a regular text string to depict a bounding box utilizes a total of $2+4\times5+3=25$ tokens, encompassing 2 trigger marks, 4 floating-point numbers, and 3 commas. However, by substituting multiple digit tokens of each floating-point number with a unique coordinate token and remaining only 1 comma, we can lower the number of tokens to just $2+4+1=7$.

However, solely training the newly appended word embeddings with language modeling loss on a small amount of data is not effective. In our experiments, the model occasionally collapses, producing meaningless coordinates. To alleviate the problem of inefficient training of coordinate tokens, we aim to introduce an auxiliary training target. Taking inspiration from DETR~\cite{detr}, we incorporate a straightforward 2-layer MLP with ReLU activation function and a linear projection layer as the auxiliary detection head, which runs in parallel with the original output layer of the LLM. The output of the detection head is normalized by the Sigmoid activation function. We evaluate the error between the prediction and the ground truth by $\ell_1$ loss:
\begin{align}
    \calL_\text{box}&=\frac{1}{|\mathcal{B}|}\sum_{i\in \calB}\|b_i-b^*_i\|_1,
\end{align}
where $b_i$ and $b^*_i$ are the predictions and the ground truth of normalized bounding box coordinates at position $i$ respectively, and $\mathcal{B}$ is the set of coordinate token positions in the output sequence.

\subsubsection{Loss Function.} All of the data is organized into multi-turn conversations, with each turn formatted as:
\begin{align}
    \text{User: <s>}\calI^t\text{</s>Assistant: <s>}\calR^t\text{</s>}
\end{align}
where <s> and </s> are special tokens denoting the beginning and end of conversation messages. $\mathcal{I}^t$ and $\mathcal{R}^t$ are the instruction tokens and response tokens at the $t$-th turn. Unlike language instruction tuning which only involves text tokens, $\mathcal{I}^t$ might consist of text, image, or both modality tokens. The training of MLLMs is mainly based on the language modeling loss over the response tokens:
\begin{eqnarray}
    \calL_\text{lm}=-\frac{1}{\sum \alpha_i}\sum_{i\in \calM}\alpha_i\log(p(x_i|\varx_{<i})),\quad
    \alpha_i=\left\{
    \begin{aligned}
        &1\quad&i\notin\calB,\\
        &\alpha&i\in\calB,
    \end{aligned}
    \right.
\end{eqnarray}
where $\calM$ is the set of response positions, $\alpha$ is a predefined weight for coordinate tokens, and $\varx_{<i}$ are multimodal context-containing instruction and response tokens that appeared before the $i$-th token.

The final loss is the weighted sum of the language modeling loss and the aforementioned bounding box loss:
\begin{align}
    \calL=\calL_\text{lm} + \lambda\calL_\text{box},
\end{align}
where $\lambda$ is a predefined weight for the bounding box loss.

\section{Experiments}

\begin{figure*}
    \centering
    \includegraphics[width=\textwidth]{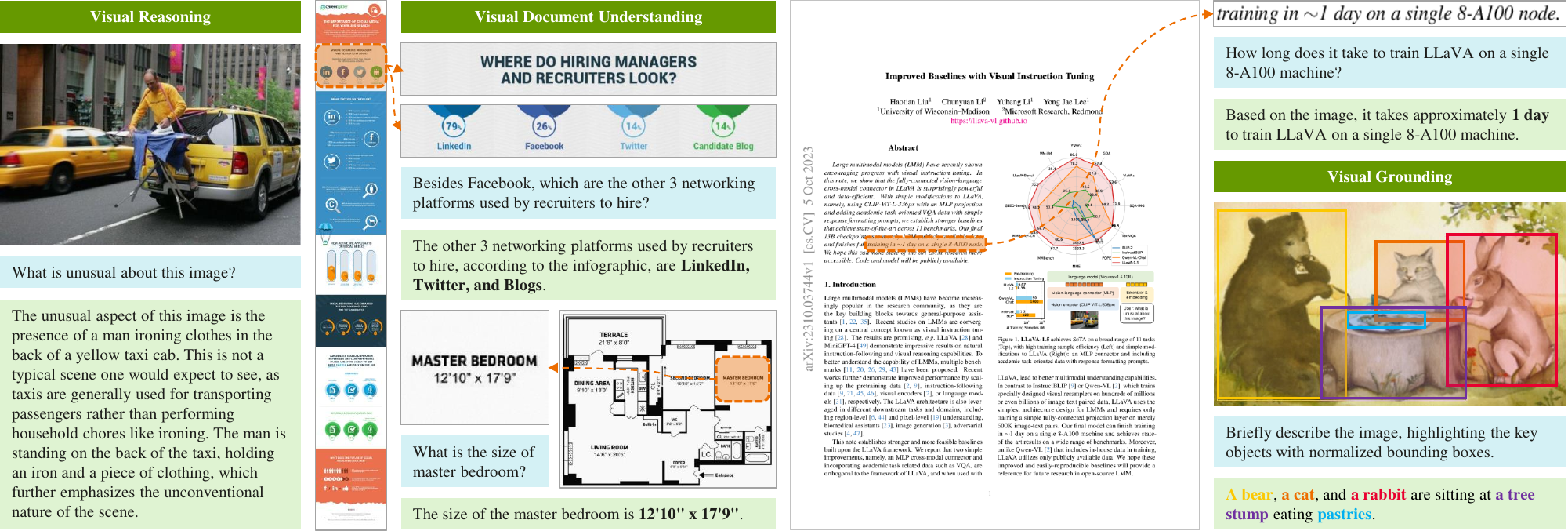}
    \caption{Examples of the results produced by TextHawk.}
    \label{fig:visu}
\end{figure*}

\subsection{Datasets}

\subsubsection{Data Concatenation.} Creating data batches with sequences of varying lengths requires padding, resulting in the waste of tokens. To mitigate this inefficiency and increase training throughput, we combine multiple native samples into a single training sample. Specifically, we select and concatenate samples from the dataset randomly until the combined sequence length reaches a predefined maximum value. It is worth noting that we carefully mask the native samples so that they are \textit{mutually invisible} from each other.

\subsubsection{Conceptual Captioning.} To bridge basic perception capability as well as align concept between visual encoder and LLM, 96M image-text pairs are collected from image captioning datasets, including CC3M~\cite{cc3m}, CC12M~\cite{cc12m}, SBU~\cite{sbu} and a subset of LAION-400M~\cite{laion400m}. In this task, the model generates a short caption for the given image, as required by the prompt "\textit{Briefly describe the image}".

\subsubsection{Grounding Captioning.} To empower MLLM with basic grounding capability, a subset of GrIT~\cite{kosmos2} including 16M image-text pairs is adopted. In this task, the model generates a short caption as well as the normalized bounding boxes of referring objects in the image, as required by the prompt "\textit{Briefly describe the image, highlighting the key objects with normalized bounding boxes}".

\subsubsection{OCR.} Except for natural images, we are particularly interested in document-oriented images. To enhance the perception capability of MLLM for optical characters, 1.28M images from IIT-CDIP~\cite{iit_cdip} are collected. Three kinds of queries, "\textit{List the text content in the image}", "\textit{List the text bounding boxes in the image}" and "\textit{List the text content along with their bounding boxes in the image}", are used to prompt the model to generate the text content, bounding boxes, or both of them for a given image, of which the coarse labels are collected with a commercial OCR system.

\subsubsection{Markdown.} Inspired by Nougat~\cite{nougat}, we collect 1.28M PDF pages and corresponding Markdown content of scientific papers from arXiv source files, which contain more layout information such as reading order than regular OCR data. We use a simple instruction, "\textit{Transcribe the content of the document image}", to ask the model to convert a PDF page of scientific paper into Markdown.

\subsubsection{Instruction.} Following LLaVA-1.5, we build our fine-tuning data based on existing datasets to enhance the instruction-following and chatting capability of MLLMs on nature and document-oriented tasks. Specifically, we adopt multiple datasets including VQAv2~\cite{vqav2}, OK-VQA~\cite{okvqa}, GQA~\cite{gqa}, A-OKVQA~\cite{aokvqa}, TextCaps~\cite{textcaps}, OCR-VQA~\cite{ocrvqa}, RefCOCO~\cite{refcoco}, PointQA~\cite{pointqa}, Flickr~\cite{flickr}, DocVQA~\cite{docvqa}, ChartQA~\cite{chartqa}, InfographicVQA~(InfoVQA)~\cite{infovqa}, TabFact~\cite{tabfact}, WikiTableQuestions~(WTQ)~\cite{wtq}, VG~\cite{vg}, VisualMRC~\cite{visualmrc}, and SlideVQA~\cite{slidevqa}.
The same prompts from LLaVA-1.5 are adopted to regularize the response style of MLLMs.
For each dataset, we concatenate all of the QA pairs corresponding to the same training image to create multi-turn conversations and improve data efficiency. Except for the original tasks, we additionally introduce multiple tasks to help the MLLMs recognize text and understand document layout, including OCR task for DocVQA, InfoVQA, VisualMRC and SlideVQA, chart-to-table task for ChartQA, and image-to-markdown task for TabFact and WTQ. To develop a MLLM for general purpose, we make use of several dialogue datasets including ShareGPT, ShareGPT-4V~\cite{sharegpt4v}, ALLaVA~\cite{allava}, LLaVA~\cite{llava}, SVIT~\cite{svit}, and Shikra~\cite{shikra}.

\subsubsection{DocGemini.} To address the scarcity of high-quality document-oriented dialogue datasets, we leverage the native visual capabilities of Gemini-Pro for data augmentation. For each training sample from DocVQA, ChartQA, and InfoVQA, we provide Gemini-Pro the image and original QA pairs along with a query for generating: (1) a brief summary of the document topics; (2) extra short QA pairs, up to 10; (3) insights behind each answer. In summary, the generated dataset \textit{DocGemini} consists of 30K images and 195K QA pairs with insights.

\begin{table*}[tb]
  \caption{Results on vision-language benchmarks. $\textbf{\modelname}^\dag$ is fine-tuned without the DocGemini. The values in bold and underlined are the best and second-best results.}
  \label{tab:benchmark}
  \centering
  \setlength{\tabcolsep}{0.8mm}\begin{tabular}{l|c|cccc|ccccc|ccc}
    \toprule
    \multirow{2}{*}{Model} & \multirow{2}{*}{ViT~(Params.)}  & \multicolumn{4}{c|}{General} & \multicolumn{5}{c|}{Document} & \multicolumn{3}{c}{RefCOCO}\\
    & & $\text{MME}^\text{P}$ & $\text{MMB}^\text{dev}$ & $\text{SEED}^\text{I}$ & GQA & DocVQA & ChartQA & InfoVQA & TabFact & WTQ & val & test-A & test-B\\
    \hline
    \rowcolor{lightgray!50}\multicolumn{14}{l}{\textcolor{gray}{Specialist}}\\
    Donut~\cite{donut} & Swin-B~(0.1B) & - & - & - & - & 67.5 & 41.8 & 11.6 & 54.6 & 18.8 & - & - & -\\
    Pix2Struct~\cite{pix2struct} & - & - & - & - & - & \textbf{76.6} & 58.6 & 40.0 & - & - & - & - & -\\
    \rowcolor{lightgray!50}\multicolumn{14}{l}{\textcolor{gray}{Generalist}}\\
    InternLM-XC~\cite{xcomposer} & EVA-G~(1B) & \textbf{1528.4} & \textbf{74.8} & 66.1 & - & - & - & - & - & - & - & - & -\\
    LLaVA-1.5-7B~\cite{llava-1.5} & CLIP-L~(0.3B) & 1510.7& 65.2 & - & 62.0 & - & - & - & - & - & - & - & -\\
    Shikra-7B~\cite{shikra} & CLIP-L~(0.3B) & - & 58.8 & - & - & - & - & - & - & - & 87.0 & \underline{91.1} & 81.8\\
    Qwen-VL-Chat~\cite{qwen-vl} & CLIP-G~(2B) & 1487.6 & 60.6 & 65.4 & 57.5 & 62.6 & 66.3 & - & - & - & \textbf{88.6} & \textbf{92.3} & \textbf{84.5}\\
    Monkey~\cite{monkey} & CLIP-G~(2B) & - & 59.3 & - & 60.7 & 66.5 & 65.1 & 36.1 & - & 25.3 & - & - & -\\
    UReader~\cite{ureader} & CLIP-L~(0.3B) & - & - & - & - & 65.4 & 59.3 & 42.2 & 67.6 & 29.4 & - & - & -\\
    TextMonkey~\cite{ureader} & CLIP-G~(2B) & - & - & - & - & 73.0 & \textbf{66.9} & - & - & 31.9 & - & - & -\\
    $\textbf{\modelname}^\dag$ & \text{SigLIP-SO}~(0.4B) & \underline{1520.9} & 73.0 & \textbf{69.2} & \textbf{64.7} & \underline{73.6} & 64.0 & \underline{47.3} & \underline{70.7} & \underline{33.5} & \underline{87.3} & 90.9 & \underline{83.3}\\
    \textbf{\modelname} & \text{SigLIP-SO}~(0.4B) & 1500.0 & \underline{74.6} & \textbf{69.2} & \underline{64.6} & \textbf{76.4} & \underline{66.6} & \textbf{50.6} & \textbf{71.1} & \textbf{34.7} & 87.2 & 90.8 & 82.5\\
  \bottomrule
  \end{tabular}
\end{table*}

\subsection{Training}
\label{sec:train}

For all of the training stages, we adopt AdamW as the optimizer, with $\beta_1=0.9$, $\beta_2=0.95$, and a weight decay of 0.05.

\subsubsection{Fixed Resolution Pre-Training.} Inspired by BLIP-2, we adopt large-scale conceptual captioning datasets to align a pre-trained and frozen visual encoder with LLM. Specifically, 96M image-text pairs are used in this stage. Each conceptual caption is a brief description summarizing the overall information portrayed in an image, rarely related to the fine-grained details. To accelerate training, all images undergo resizing to $224\times224$. The maximum sequence length is 4,096 and the batch size is 96, resulting in an effective batch size of nearly 8,000 after data concatenation. We pre-train the model for 12,000 steps, equivalent to almost 1 epoch across the entire dataset. During pre-training, we freeze the visual encoder and LLM and train the randomly initialized resampler and LoRA modules. The learning rate linearly warmup to $3e^{-4}$ in the first 3\% steps, followed by cosine decay to $1e^{-5}$ in the remaining steps. It takes 1 day to finish training on 48 NVIDIA V100 GPUs.

\subsubsection{Mixed Resolution Pre-Training.} In this stage, we adapt the resampler to variable resolution input. Images with different native sizes and aspect ratios from the grounding captioning, OCR, and Markdown datasets are used. The size of each sub-image is $224\times224$. The maximum area $n$ is set to $36$ and the maximum side-length $l$ is set to $12$. To accelerate the grid matching for shape-adaptive cropping, $k$ is set to 9. The effective batch size is nearly 1,500 and the number of training steps is 12,000, equivalent to almost 1 epoch across the entire dataset. Except for the resampler and LoRA, a detection head is randomly initialized and updated in this stage. The weight $\alpha$ for coordinate tokens is set to $0.25$~(4 tokens per bounding box) and the weight $\lambda$ for $\ell_1$ loss is set to 1. The visual encoder and LLM are kept frozen. The learning rate linearly warmup to $1.5e^{-4}$ in the first 3\% steps, followed by cosine decay to $5e^{-6}$. It takes 3 days to finish training on 40 NVIDIA V100 GPUs.

\subsubsection{Mixed Resolution Supervised Fine-Tuning.} During fine-tuning, we merge LoRA weights into LLM and seamlessly train the resampler, LLM, and detection head together, while keeping the visual encoder frozen. The hyper-parameters for the shape-adaptive cropping and the detection head are inherited from mixed resolution pre-training. The maximum sequence length is 2,048. We train the model on instruction-following data for 1 epoch with a batch size of 64. The learning rate linearly warmup to $2e^{-5}$ in the first 3\% steps, followed by cosine decay to $0$. It takes 1 day to finish training on 32 NVIDIA V100 GPUs.

\subsection{Results on Standard Benchmarks}
To demonstrate the effectiveness of our methods, we conduct a comparison among \modelname, two specialist models for document-oriented tasks, and recent MLLMs on a wide range of benchmarks. Some qualitative results are shown in Fig.~\ref{fig:visu}. Each benchmark targets a group of general-purpose tasks or fined-grained tasks. Firstly, we evaluate the models on comprehensive benchmarks including MME~\cite{mme}, MMBench~\cite{mmb}, SEED-Bench~\cite{seedbench}, and GQA~\cite{gqa}. Since the image resolutions of these benchmarks are relatively low, we further evaluate the capability of fined-grained perception on document understanding and referring tasks, including DocVQA~\cite{docvqa}, ChartQA~\cite{chartqa}, InfoVQA~\cite{infovqa}, TabFact~\cite{tabfact}, WTQ~\cite{wtq}, and RefCOCO~\cite{refcoco}.

As depicted in Table~\ref{tab:benchmark}, \modelname excels in both general and document-oriented benchmarks, securing the top spot in 6 out of 9 benchmarks. In all the general benchmarks, \modelname not only surpasses LLaVA-1.5-7B~\cite{llava-1.5}, but also achieves comparable results with InternLM-XComposer~\cite{xcomposer}, despite the latter sharing the same foundational LLM but utilizing a larger visual encoder. When compared to previous document-oriented MLLMs, such as Ureader~\cite{ureader} and TextMonkey~\cite{textmonkey}, \modelname demonstrates superior performance on document-oriented benchmarks. Specifically, \modelname achieves performance gains of $11.0\%$, $7.3\%$, $8.4\%$, $3.5\%$, and $5.3\%$ on DocVQA, ChartQA, InfoVQA, TabFact, and WTQ, respectively, when compared to Ureader. Remarkably, \modelname even surpasses TextMonkey, which employs a larger visual encoder, on DocVQA and WTQ benchmarks. It is worth mentioning that the introduction of our DocGemini data can further improve the performance on the document-oriented benchmarks.
Besides, \modelname achieves competing results on the RefCOCO dataset, showing its good capabilities on the referring task.

\subsection{Ablation Study}

We adopt two faster training configurations for the ablation study. The fixed resolution pre-training is exactly the same as what is described in Sec~\ref{sec:train}. Subsequently, fixed resolution models are fine-tuned only on the training data of LLaVA-1.5 for 1 epoch, while variable resolution models are fine-tuned on the training data of LLaVA-1.5, DocVQA, ChartQA, InfoVQA, TabFact, and WTQ.

\begin{table}[tb]
  \caption{Effect of combining resampling and rearrangement for compressing visual tokens progressively.}
  \label{tab:resa}
  \centering
  \setlength{\tabcolsep}{2mm}\begin{tabular}{cc|cccc}
    \toprule
    Resample & Rearrange & $\text{MMB}^\text{dev}$ & GQA & $\text{RefCOCO}^\text{val}$\\
    \midrule
    $256 \rightarrow 16$ & -                    & 72.19 & 60.05 & 50.61\\
    $256 \rightarrow 64$ & $64 \rightarrow 16$  & 72.45 & 60.38 & 52.78\\
  \bottomrule
  \end{tabular}
\end{table}

\subsubsection{ReSampling and ReArrangement (ReSA).} To demonstrate the effectiveness of ReSA, we conduct fixed resolution experiments with different compression configurations, and the results are shown in Table~\ref{tab:resa}. Compared to the resampling-only strategy, incorporating ReSA which divides the compression procedure into two stages improves performance on all benchmarks, especially on RefCOCO as the referring comprehension task exhibits a great demand for preserving more fine-grained information.

\begin{table}[tb]
  \caption{Comparison of different routing tables for multi-level cross-attention. Numbers in brackets represent from which stage the features are extracted.}
  \label{tab:mlca}
  \centering
  \setlength{\tabcolsep}{2mm}\begin{tabular}{c|c|cccc}
    \toprule
    \# & MLCA & $\text{MMB}^\text{dev}$ & GQA & $\text{RefCOCO}^\text{val}$\\
    \midrule
    1 & $[3,3,3,3,3,3,3,3]$ & 72.45 & 60.38 & 52.78\\
    2 & $[3,2,1,0,0,1,2,3]$ & 71.94 & 60.56 & 55.24\\
    3 & $[3,2,1,0,3,2,1,0]$ & 71.94 & 60.37 & 56.14\\
    4 & $[3,3,2,2,1,1,0,0]$ & 72.19 & 59.83 & 56.43\\
    5 & $[3,3,3,2,2,2,1,0]$ & 72.53 & 60.10 & 56.75\\
  \bottomrule
  \end{tabular}
\end{table}

\subsubsection{Multi-Level Cross-Attention (MLCA).}
Empirically, deep layers within visual encoders primarily capture global semantic information, while shallow layers tend to retain local, intricate details. To explore the effect of the routing strategy of MLCA, we conduct experiments with different routing tables, shown in Table~\ref{tab:mlca}. For the sake of simplicity, we use R1 to R5 to denote different routing tables. R1 is a special case that only includes encoder stage 3, degrading to the vanilla cross-attention settings. Comparing R1 and R2, we can find the latter significantly improves the performance on fine-grained tasks, while slightly sacrificing the performance on the general benchmarks. Comparing R2 and R3/R4, we can find routing features from shallower encoder layers to deeper resampler layers demonstrate higher accuracy on RefCOCO, compared to routing them to intermediate resampler layers. Among all experimental settings, R5 achieves a good balance between general tasks and fine-grained tasks, hence we adopt it as the default routing table.

\begin{table}[tb]
  \caption{Effect of incorporating query proposal network.}
  \label{tab:qpn}
  \centering
  \setlength{\tabcolsep}{2mm}\begin{tabular}{c|ccccc}
    \toprule
    QPN & $\text{MME}^\text{P}$ & $\text{MMB}^\text{dev}$ & GQA & $\text{RefCOCO}^\text{val}$\\
    \midrule
               & 1471.3 & 72.53 & 60.10 & 56.75\\
    \checkmark & 1507.2 & 72.02 & 61.07 & 58.21\\
  \bottomrule
  \end{tabular}
\end{table}

\subsubsection{Query Proposal Network (QPN).} To validate the importance of high-quality resampler queries, we compare initializing queries from learned parameters and generating queries with QPN, as shown in Table~\ref{tab:qpn}. For a fair comparison, the number of queries is 64 in both experiments. We can find incorporating QPN improves model performance on most benchmarks, especially on RefCOCO.

\begin{table}[tb]
  \caption{Effect of incorporating positional embeddings, where \textit{APEs} denotes absolute positional embeddings, and \textit{SPEs} denotes scalable positional embeddings. In the field of granularity, \textit{cell} and \textit{patch} mean applying different embeddings for each sub-image and patch respectively.}
  \label{tab:pe}
  \centering
  \setlength{\tabcolsep}{1mm}\begin{tabular}{cc|cccc}
    \toprule
    PE & Granularity & $\text{RefCOCO}^\text{val}$ & DocVQA & ChartQA & InfoVQA\\
    \midrule
    -   & -           & 79.13 & 67.68 & 61.04 & 39.77\\
    APEs & cell        & 82.03 & 68.55 & 61.02 & 43.28\\
    SPEs & cell        & 82.65 & 69.63 & 61.32 & 43.03\\
    SPEs & patch       & 83.74 & 69.65 & 61.96 & 43.85\\
  \bottomrule
  \end{tabular}
\end{table}

\subsubsection{Scalable Positional Embeddings (SPEs).} To explore the effect of additional positional embeddings, we conduct experiments with variable resolution settings. The results on fine-grained benchmarks are shown in Table~\ref{tab:pe}. Apparently, the absence of additional positional embeddings leads to performance degradation on most benchmarks. Compared with absolute positional embeddings used in previous works, SPEs further improve fine-grained performance. Meanwhile, the granularity of SPEs can be extended from cell to patch without increasing the number of parameters. It is confirmed that using finer and smoother positional embeddings at the image patch level further improves the overall performance.

\begin{table}[tb]
  \caption{Comparison of heads on decoding coordinates.}
  \label{tab:head}
  \centering
  \setlength{\tabcolsep}{1mm}\begin{tabular}{l|ccc}
    \toprule
    Head & $\text{RefCOCO}^\text{val}$ & $\text{RefCOCO}^\text{test-A}$ & $\text{RefCOCO}^\text{test-B}$\\
    \midrule
    Language & 85.6 & 90.2 & 80.6\\
    Detection & 87.3 & 90.9 & 83.3\\
  \bottomrule
  \end{tabular}
\end{table}

\subsubsection{Detection Head.} Both the original language modeling head and the additional detection head are capable of generating coordinates. Whenever the former produces a coordinate token, we can seamlessly substitute it with the output from the latter. In Table~\ref{tab:head}, we compare the results of different heads on RefCOCO. It is obvious that the detection head demonstrates higher accuracy on all splits, proving its superiority on the grounding tasks.

\section{Limitations}
The visual encoder in \modelname is frozen during training, which means it does not learn from the training data. This could limit the model’s ability to adapt to new or unseen visual data that significantly differs from the data it was initially trained on. In the future, we will train the vision encoder to further improve the perception capabilities.

\section{Conclusion}
In this paper, we have presented \modelname, a novel Multimodal Large Language Model (MLLM) that is specifically designed to address the unique challenges posed by document-oriented tasks. \modelname introduces several innovative components. These components work in synergy to enhance the model’s fine-grained visual perception and information compression capabilities, thereby enabling it to handle the high resolution and information density characteristic of document images. Our extensive experiments on both document-oriented and general MLLM benchmarks demonstrate that \modelname outperforms state-of-the-art methods, showcasing its superior fine-grained document perception and general vision-language abilities.


\bibliographystyle{ACM-Reference-Format}
\bibliography{sample-base}

\end{document}